\documentclass[sigconf]{acmart}
\renewcommand\footnotetextcopyrightpermission[1]{}
\setcopyright{none}
\usepackage{blindtext}
\usepackage{enumitem}
\usepackage{xcolor}
\usepackage{threeparttable}
\usepackage{cleveref}
\usepackage{multirow}
\usepackage{booktabs}
\usepackage{makecell}
\usepackage{bm}

\begin{document}

\title{A Two-Phase Approach for Abstractive Podcast Summarization}

\author{Chujie Zheng}
\email{chz@udel.edu}
\affiliation{%
  \institution{University of Delaware, USA}
}

\author{Kunpeng Zhang}
\email{kpzhang@umd.edu}
\affiliation{%
  \institution{University of Maryland, USA}
}

\author{Harry Jiannan Wang}
\email{hjwang@udel.edu}
\affiliation{%
  \institution{University of Delaware, USA}
}

\author{Ling Fan}
\email{lfan@tongji.edu.cn}
\affiliation{%
  \institution{Tongji University, China}
}

\renewcommand{\shortauthors}{Zheng et al.}

\begin{abstract}
Podcast summarization is different from summarization of other data formats, such as news, patents, and scientific papers in that podcasts are often longer, conversational, colloquial, and full of sponsorship and advertising information, which imposes great challenges for existing models. In this paper, we focus on abstractive podcast summarization and propose a two-phase approach: sentence selection and seq2seq learning. Specifically, we first select important sentences from the noisy long podcast transcripts. The selection is based on sentence similarity to the reference to reduce the redundancy and the associated latent topics to preserve semantics. Then the selected sentences are fed into a pre-trained encoder-decoder framework for the summary generation. Our approach achieves promising results regarding both ROUGE-based measures and human evaluations.
\end{abstract}

\maketitle

\pagestyle{plain}

\section{Introduction}
The summarization task has been well studied in Natural Language Processing (NLP). Especially within the development of deep learning and the introduction of attention mechanism\cite{vaswani2017attention}, Transformer-based summarization models \cite{yan2020prophetnet, raffel2019exploring, lewis2019bart, zhang2019pegasus} have achieved remarkable performance. However, these Transformer-based models are unable to process long sequences due to their self-attention operation, which scales quadratically with the sequence length \cite{beltagy2020longformer, kitaev2020reformer}. This bottleneck has brought challenges for podcast summarization since the average length of podcast transcript is much longer than the maximum sequence limitation. When these Transformer-based summarization models can only read the first hundreds of tokens in the episode transcript, how to generate a comprehensive summary covering the important information is a challenge.

Besides the challenge from the length, podcast summarization is a complicated research problem because of the conversational feature. Comparing to the professionally edited texts like news and academic papers, the podcast is colloquial and contains many conversations. According to the number from Spotify, 81.4\% of the podcast episodes are conversational\footnote{Numbers are reported in Matthew Sharpe's presentation "A Review of Metadata Fields Associated with Podcast RSS Feeds" in RecSys 2020 PodRecs Workshop}. But few studies have focused on how to deal with conversational corpus summarization. It is still a research gap.

\begin{figure*}
\centering
\includegraphics[width=0.8\textwidth]{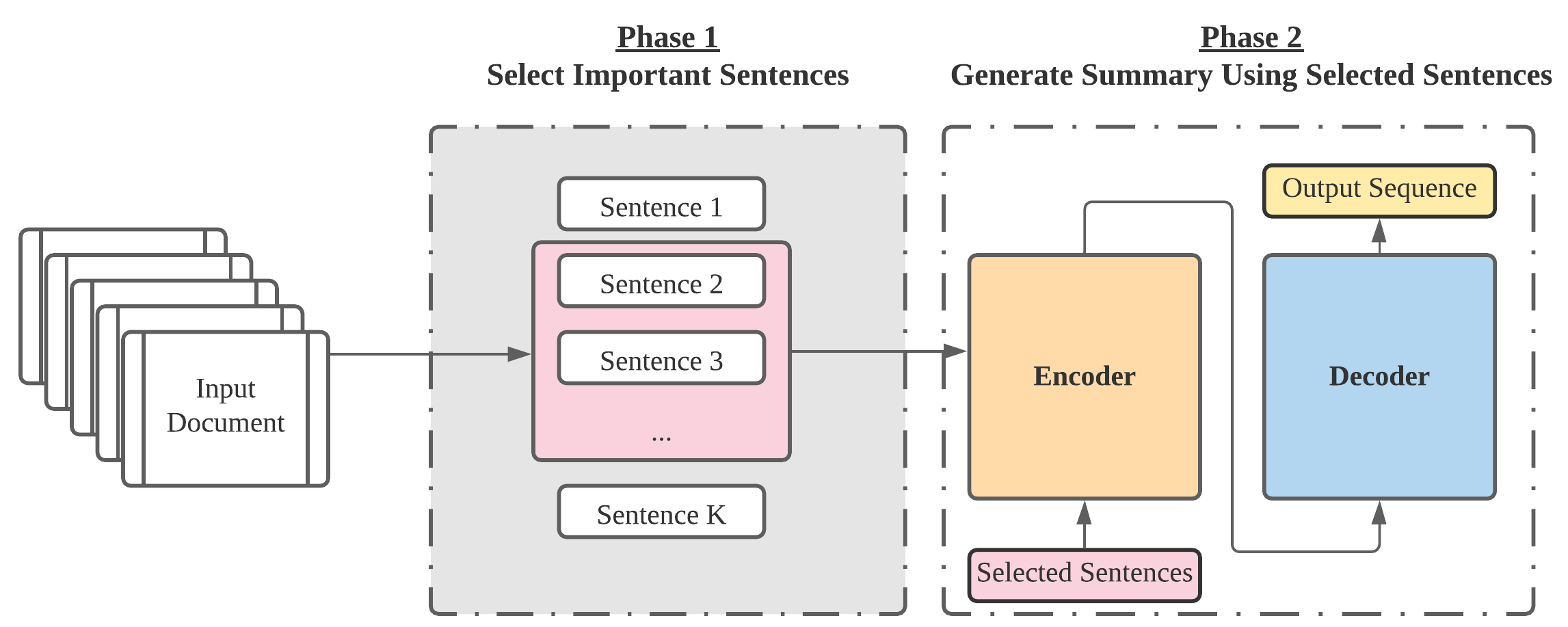}
\caption{Framework for Two-Phase Approach}
\label{fig::framework-two-phase}
\end{figure*}

This short paper is for the summarization task of TREC 2020 Podcast Track \cite{trec2020podcastnotebook}. In this work, we introduce a two-phase approach for podcast abstractive summarization. It is designed based on the unique features of the podcast. Our proposed approach selects the important sentences from the transcript in the first phase and uses the encoder-decoder network to generate the abstractive summary based on the selection. Figure \ref{fig::framework-two-phase} describes the general framework for proposed approach.

The key contribution of this paper focuses on the first phase, which selects the important sentences from the input documents. Building on top of the Transformer-based models, our goal is to filter the irrelevant content and locate the most useful information for the abstractive summary generation. Under this idea, the research question becomes how to define important sentences? We propose two methods to identify the important sentences using the ROUGE score and the topic-level features. Our proposed approach provides some novel insights for podcast summarization and improves the performance of the summarization model.  


\section{Dataset}
The dataset used in this work is the TREC Spotify podcast dataset \cite{clifton2020spotify, clifton2020hundredthousand} which has 105,360 podcast episodes from 18,376 shows produced by 17,473 creators. The average duration of a single episode is 30 minutes, while the longest can be over 5 hours and the shortest is only 10 seconds. The TREC Podcast Track organizers form the "Brass Set" by cutting down the dataset to 66,245 podcast episodes using the following rules:

\begin{table*}
\centering
\begin{tabular}{l|c}
\hline
\multicolumn{1}{c|}{\textbf{Dataset Preprocessing}} & \multicolumn{1}{c}{\textbf{\# of Episodes}} \\
\hline
TREC Spotify Podcasts Dataset & 105,360\\
After filtering by the TREC organizer (Brass Set)  & 66,245 \\
After removing episodes with profanity language & 55,799 \\
After removing episodes with non-English descriptions & 55,383 \\
After removing episodes whose description is too short & 52,140 \\
\hline
\end{tabular}
\caption{Data Preprocessing and the Number of Episodes}
\label{data::preprocessing}
\end{table*}

\begin{itemize}
    \item Remove episodes with descriptions that are too long (> 750 characters) or too short (< 20 characters); 
    \item Remove "duplicate" episodes with similar descriptions (by conducting similarity analysis); 
    \item Remove episodes with descriptions that are similar to the corresponding show descriptions, which means the episode description may not reflect the episode content. 
\end{itemize}

We perform additional data preprocessing on top of the Brass Set as follows:

\begin{itemize}
    \item Remove episodes with profanity language in the episode or show descriptions as \cite{raffel2019exploring}.
    \item Remove episodes with non-English descriptions.
    \item Remove episodes whose description is less than 10 tokens\footnote{We perform some data preprocessing for episode description, including using some rule-based methods to remove the social media link and sponsorship.}.
\end{itemize}

After preprocessing, the dataset has 52,140 episodes left (see Table \ref{data::preprocessing} for details). We randomly split the dataset into training, validation, and testing by 80\%, 10\%, and 10\%. This processed dataset is used for our training.

\section{Two-Phase Approach}
In this section, we introduce our \textbf{Two-Phase Approach} for podcast summarization. The key motivation is to select important sentences from the input document and use it as the input for the abstractive summarization model. In this work, we don't design a new model for Phase 2, so we choose to use the BART framework, which performs the best in the preliminary experiment \cite{zheng2020baseline}. In practical implementation, we use distilBART \footnote{https://huggingface.co/sshleifer/distilbart-cnn-12-6} provided by HuggingFace. Our focus is on how to define an efficient metric to select sentences contained important information. So the research problem turns to how to define the important sentences.

We propose two sentence selection methods to identify the sentences with important information. The importance is determined based on whether the sentence covers the key information of the article or whether it reflects the key topic of the article.


\begin{figure*}
\centering
\includegraphics[width=0.6\textwidth]{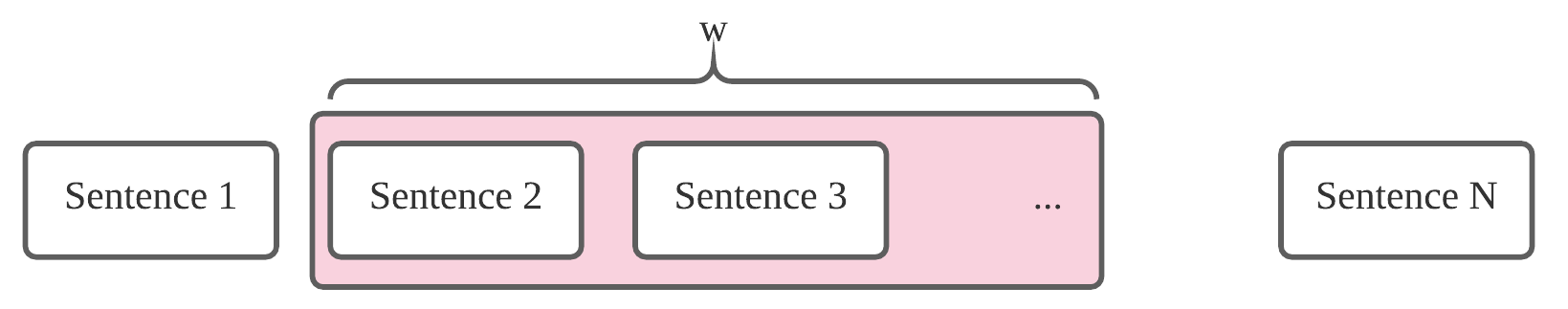}
\caption{Framework for ROUGE-based Approach}
\label{fig::framework-rouge-based}
\end{figure*}

\begin{figure*}
\centering
\includegraphics[width=0.7\textwidth]{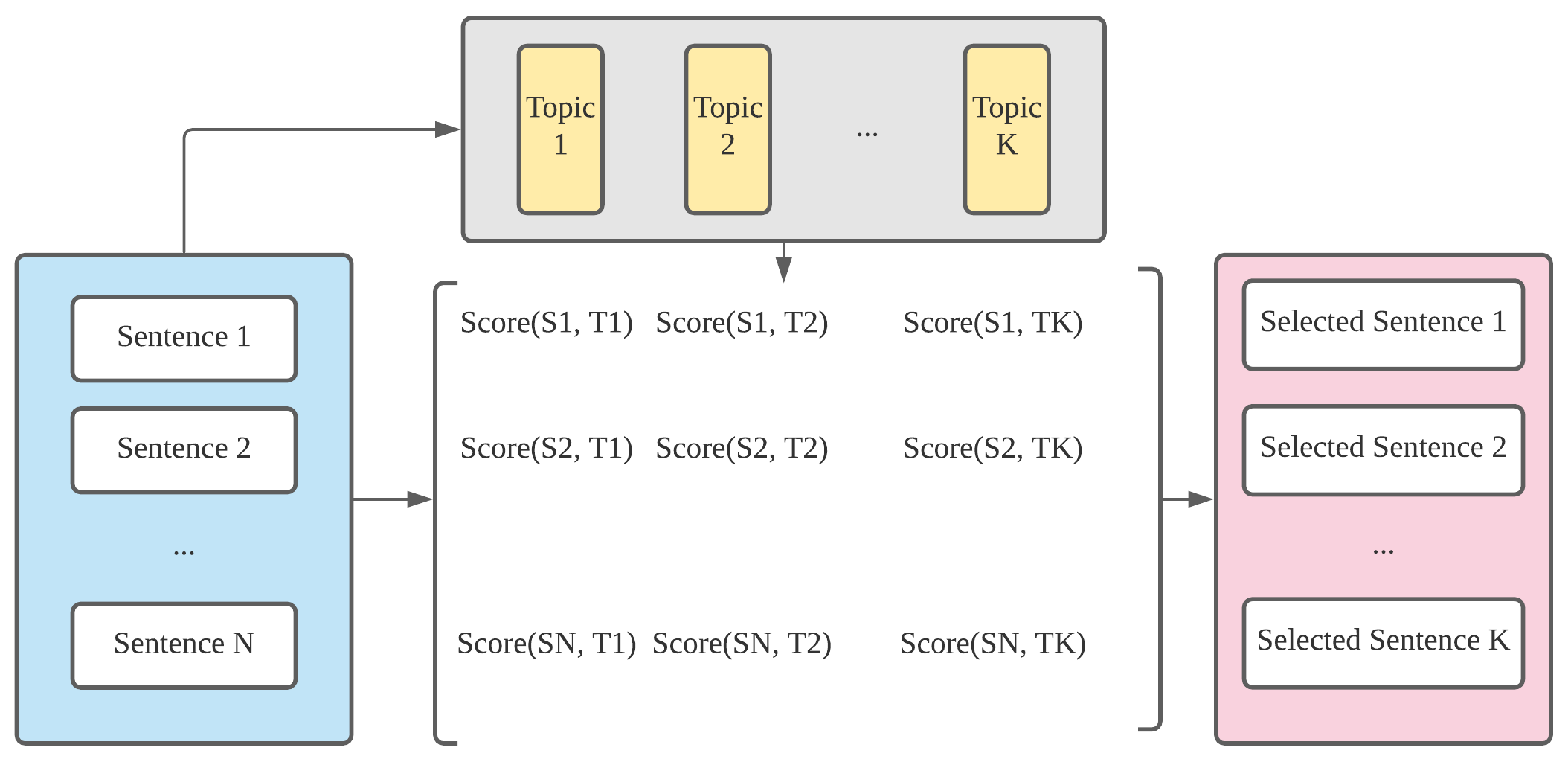}
\caption{Framework for Topic-Enhanced Approach}
\label{fig::framework-topic-enhanced}
\end{figure*}

\subsection{ROUGE-based Approach}
Inspired by \cite{zhang2019pegasus}, we propose a ROUGE-based approach to identify the sentence who covers the key information of the input document. We calculate the ROUGE score \cite{lin2004rouge} for each sentence with the input document, which considers as the relevance score for this sentence. ROUGE is the most common evaluation metric for the summarization task, which focuses on the overlapping between the reference summary and the model output. The intuition for the \textbf{ROUGE-based Approach} is a sentence with a higher ROUGE score serves as an excellent summary of the input document, which should capture the important information. 

In practical implementation, since there are three types of ROUGE scores commonly used: ROUGE-1, ROUGE-2, and ROUGE-L, we choose the average of three scores as the measurement to find the most relevant sentences. We introduce our detailed implementation \textbf{Sliding Window ROUGE-based Approach} and \textbf{Novelty-Enhanced ROUGE-based Approach} in the following.

\subsubsection{Sliding Window ROUGE-based Approach}
We select the sentences in a window size as the candidate and compute the ROUGE score, as Figure \ref{fig::framework-rouge-based}. This implementation is based on two reasons. First, it is very time consuming to calculate the score for every single sentences, especially for the long document. In addition, we believe selecting single sentence may lead to the redundancy in the information. In other words, we are using ROUGE score as the measurement to selecting sentences \textbf{x} in the window size $w$, as Equation \ref{eq::rouge-based}. $s_k$ represents the $k$-th sentence in the input document, which contains $N$ sentences.
\begin{equation}
	\textbf{x} = \text{argmax}_{s_{i:i+w}}\text{ROUGE}(s_{i:i+w}, s_{1:N})
	\label{eq::rouge-based}
\end{equation}

Here $w$ is chosen based on the consideration that the average length of the selected sentences should have approximately 1024 tokens, which is the maximum sequence limit we set for BART, followed the instructions provided by HuggingFace \footnote{https://github.com/huggingface/transformers/tree/master/examples/seq2seq}. In our implementation, we set $w=40$ since the average number of tokens is 1018 tokens, which satisfied our constraints. 

\subsubsection{Novelty-Enhanced ROUGE-based Approach}
From the implementation of the Sliding Window ROUGE-based Approach, it brings us another question: since we are selecting the sentences in consecutive order, is it possible that we may miss some single important sentences? To solve this question, we propose a new approach: besides finding the sentences $s_{i:i+w}$ in the window size $w$, we also calculate the ROUGE score for each sentence $s_j$ and check if it is selected already in the window. If not, we will combine it with the selected window as the summary. This approach aims to enhance the performance of the ROUGE-based approach in case we miss any important single sentence.

We choose the top $k=5$ sentences with the highest ROUGE score and check if these sentences are in the selected sentences under the window size $w=25$. These two parameters are chosen based on the requirement that the average length of the final selection satisfies the maximum length limit for the Transformer.

\subsection{Topic-Enhanced Approach}
The second proposed approach for important sentence selection is identifying the sentence captures the main topic of the input document. The intuition is to learn the main topics at first and find the sentences that are most relevant to the key topics. 

This approach is built on the assumption for topic modeling: documents are represented as random mixtures over topics and each topic is characterized by a distribution over words \cite{blei2003latent}. As Figure \ref{fig::framework-topic-enhanced} describes, we identify the latent topics and calculate the relevance score for each sentence with the topic embedding. The sentences with the highest relevance score are selected. Using topic modeling, we generate the key topics from the input document. We use Gibbs sampling to learn the topic distribution $\bm{\theta}=(\bm{\theta}_1, \bm{\theta}_2, ..., \bm{\theta}_k)$. For each latent topic $\bm{\theta}_i$, we calculate the relevance score between each sentence and $\bm{\theta}_i$, then we select the more relevant sentence as the representation for this topic. These selected sentences are combined as the output in the first phase.

\section{Evaluation}
In this section, we summarize the evaluation result for our submitted results. Totally we've submitted 4 runs this year. The detail for each submitted run is summarized here:
\begin{itemize}
	\item \textbf{Method 1:} Using the BART framework without a sentence selection approach to fine-tune the model and generate the results.
	\item \textbf{Method 2:} Using the BART framework with sentences selected by \textbf{Sliding Window ROUGE-based Approach} to fine-tune the model and generate the results.
	\item \textbf{Method 3:} Using the BART framework with sentences selected by \textbf{Novelty-Enhanced ROUGE-based Approach} to fine-tune the model and generate the results.
	\item \textbf{Method 4:} Using the BART framework with sentences selected by \textbf{Topic-Enhanced Approach} to fine-tune the model and generate the results.
\end{itemize}

\subsection{ROUGE Score}
The organizers reported the ROUGE-L score between the generated summary and the episode description, which is written by creators. As the most common evaluation metric for summarization, the ROUGE score focuses on the overlapping between the generated summary and the reference summary. Considering the fact that the quality of the episode description varies, it may not be the perfect reference summary. For example, there are many social media links and sponsorships in the episode description, which are irrelevant content. Here the ROUGE score reported in Table \ref{table::rouge-l} is sharing to offer a standard summarization evaluation viewpoint.

\begin{table}
    \centering
    \begin{tabular}{cccc}
        \toprule
         & \textbf{ROUGE-L P} & \textbf{ROUGE-L R} & \textbf{ROUGE-L F} \\
         \hline
         \textbf{Method 1} & \underline{\textbf{23.87}} & 16.07 & \underline{\textbf{16.30}} \\
         \textbf{Method 2} & 20.84 & 16.01 & 15.29\\
         \textbf{Method 3} &  20.16 & \underline{\textbf{16.83}} & 15.49 \\
         \textbf{Method 4} & 18.44 & 13.90 & 13.29 \\
        \toprule
    \end{tabular}
    \caption{ROUGE-L Scores for Submitted Summaries}
    \label{table::rouge-l}
\end{table}

\subsection{Human Evaluation}
As one of the important parts of the evaluation, this year the organizers provide the qualitative judgments for the generated summary. They selected 179 out of 1000 episodes from the submitted set. An assessor first quickly skimmed the episode, and then made judgments for each summary for that episode, in random order. Therefore, for each episode, the assessor will give a score from 0-3 quality according to its quality, where a higher score indicates higher quality. Table \ref{table::human-evaluation-average-quality} reports the performance of our submitted results. We compare our performance with the average score for all participants, and it turns that 3 out of 4 methods perform better than the average. Compared with the average score, Method 1 with the highest quality rating scores 14.3\% higher. 

We also analyze the percentage for each category in Table \ref{table::percentage-quality-rating} for all submitted runs. Method 3 performs the best using this measurement because of its outstanding performance in the high-quality rating categories: 7.82\% of the generated summary is rated as "Excellent" and 21.79\% of the output is rated as "Good". 

\begin{table}
    \centering
	\begin{tabular}{ccccc}
	\toprule
	\textbf{Average} & \textbf{Method1} & \textbf{Method2} & \textbf{Method3} & \textbf{Method4} \\
	\hline
	0.98 & \underline{\textbf{1.12}} & 1.03 & 1.08 & 0.72\\
	\toprule	
	\end{tabular}
	\caption{Average Score for Quality Rating}
	\label{table::human-evaluation-average-quality}
\end{table}

\begin{table}
    \centering
    \begin{tabular}{ccccc}
    \toprule
    & \textbf{\%(Excellent)} & \textbf{\%(Good)} & \textbf{\%(Fair)} & \textbf{\%(Bad)} \\
    \hline
    \textbf{Method 1} & 7.26\% & \underline{\textbf{21.79\%}} & 46.37\% & 24.58\% \\
    \textbf{Method 2} & 5.59\% & 19.00\% & \underline{\textbf{48.04\%}} & 27.37\% \\
    \textbf{Method 3} & \underline{\textbf{7.82\%}} & \underline{\textbf{21.79\%}} & 41.34\% & 29.05\% \\
    \textbf{Method 4} & 3.91\% & 11.73\% & 36.87\% & \underline{\textbf{47.49\%}} \\
    \toprule
    \end{tabular}
    \caption{Percentage of Quality Rating}
    \label{table::percentage-quality-rating}
\end{table}

\begin{table}
    \centering
	\begin{tabular}{cccccc}
	\toprule
		 & \textbf{Average} & \textbf{Method1} & \textbf{Method2} & \textbf{Method3} & \textbf{Method4}\\
		\hline
		Q1 & 0.44 & \underline{\textbf{0.60}}	& 0.52 &	0.56 &	0.26   \\
		Q2 &  0.28 & \underline{\textbf{0.34}} &	0.31 &	0.28 &	0.18 \\
		Q3 & 	0.58 &  \underline{\textbf{0.70}} &	0.66 &	0.68 &	0.50  \\
		Q4 & 	0.45 & 0.51 & 	0.52 &	\underline{\textbf{0.63}} &	0.50 \\
		Q5 & 	0.52 & 0.54 & 	0.58 &	\underline{\textbf{0.62}} &	0.42  \\
		Q6 &	\underline{\textbf{0.09}} & 0.05 &	\underline{\textbf{0.09}} &	0.04 &	0.07 \\
		Q7 & 	0.62 & 0.74 & 	0.63 &	\underline{\textbf{0.75}} &	0.74  \\
		Q8 &	0.45 & 0.50 &	0.46 &	0.51 &	 \underline{\textbf{0.55}} \\
	\toprule
	\end{tabular}
	\caption{Average Score for Human Evaluation Questions}
	\label{table::human-evaluation-average-questions}
\end{table}

For every episode, the assessor is asked eight yes-or-no questions regarding the quality of the summary, and “1" indicates that the answer is “yes”. These questions include:

\begin{itemize}
	\item \textbf{Q1}: Does the summary include names of the main people (hosts, guests, characters) involved or mentioned in the podcast?
	\item \textbf{Q2}: Does the summary give any additional information about the people mentioned (such as their job titles, biographies, personal background, etc)?
	\item \textbf{Q3}: Does the summary include the main topic(s) of the podcast?
	\item \textbf{Q4}: Does the summary tell you anything about the format of the podcast; e.g. whether it's an interview, whether it's a chat between friends, a monologue, etc?
	\item \textbf{Q5}: Does the summary give you mode context on the title of the podcast?
	\item \textbf{Q6}: Does the summary contain redundant information?
	\item \textbf{Q7}: Is the summary written in good English?
	\item \textbf{Q8}: Are the start and end of the summary good sentence and paragraph start and end points?
\end{itemize}

We compare the performance of our submitted runs to the average scores for all participants. Except for Q6, a higher score represents better performance. According to Table \ref{table::human-evaluation-average-questions}, our submitted runs have achieved better performance than the average for all eight questions. Our models are better at capturing important information and semantically more fluent.

Table \ref{table::human-evaluation-average-quality}, \ref{table::percentage-quality-rating} and \ref{table::human-evaluation-average-questions} leave us a lot of room for further discussion. 
\begin{itemize}
    \item Firstly, based on the fact that the average rating is only 0.98 and the best performance for our models is 1.12, there is a lot of room for improvement for podcast summarization. Considering that the score is from 0 to 3, the average performance is only "Fair". Especially in comparison with the excellent performance of news summarization, we have a lot of thoughts: What makes podcast summarization so challenging? We are looking forward to this year's Podcast Track and improve our performance.
    \item We also raise some concerns about the prepared eight yes-or-no questions. These questions reflect the quality level of the generated summary, which is also considered as the feature for a good podcast summary. Comparing to other summaries, it emphasizes several aspects including additional information like people's job titles or the format of the podcast. These features definitely play a very important role in helping people familiar with the episode. How to capture these features leads to several directions for future research.
\end{itemize}

\section{Conclusion}
In this year's podcast track, we develop a two-phase approach to handle the podcast summarization task. Using the transcript generated by Google ASR, we design a pipeline to select the important sentences which cover the key information of the input document and generate the abstractive summary using the selected sentences. Our main contribution is that we propose two approaches to select sentences including the ROUGE-based approach and the topic-enhanced approach. These approaches provide a novel definition for important sentences and improve the performance of the podcast summarization model. 

\bibliographystyle{ACM-Reference-Format}
\bibliography{bio}

\end{document}